\documentclass[conference]{IEEEtran}
\IEEEoverridecommandlockouts

\usepackage[numbers]{natbib}
\usepackage{amsmath,amssymb,amsfonts}
\usepackage{algorithmic}
\usepackage{graphicx}
\usepackage{textcomp}
\usepackage{xcolor}
\usepackage{amssymb}
\usepackage{amsmath}
\usepackage{overpic}
\usepackage[caption=false]{subfig}
\usepackage[colorlinks,citecolor=blue]{hyperref}

\begin{document}

\title{Rapid Probabilistic Interest Learning from Domain-Specific Pairwise Image Comparisons}

\author{\IEEEauthorblockN{Michael Burke, Siyabonga Mbonambi, Purity Molala, Raesetje Sefala\thanks{Authors listed alphabetically, this work was completed during an internship at the CSIR, as part of the CSIR D-SIDE programme. M Burke was supported by funding from CSIR young researcher's establishment grant, YREF032.}}
\IEEEauthorblockA{\textit{ Mobile Intelligent Autonomous Systems} \\
\textit{Council for Scientific and Industrial Research}\\
Pretoria, South Africa \\
Contact author: michaelburke@ieee.org}
}

\maketitle

\begin{abstract}
A great deal of work aims to discover large general purpose models of image interest or memorability for visual search and information retrieval. This paper argues that image interest is often domain and user specific, and that efficient mechanisms for learning about this domain-specific image interest as quickly as possible, while limiting the amount of data-labelling required, are often more useful to end-users. This work uses pairwise image comparisons to reduce the labelling burden on these users, and introduces an image interest estimation approach that performs similarly to recent data hungry deep learning approaches trained using pairwise ranking losses. Here, we use a Gaussian process model to interpolate image interest inferred using a Bayesian ranking approach over image features extracted using a pre-trained convolutional neural network. Results show that fitting a Gaussian process in high-dimensional image feature space is not only computationally feasible, but also effective across a broad range of domains. The proposed probabilistic interest estimation approach produces image interests paired with uncertainties that can be used to identify images for which additional labelling is required and measure inference convergence, allowing for sample efficient active model training. Importantly, the probabilistic formulation allows for effective visual search and information retrieval when limited labelling data is available.
\end{abstract}



\section{Introduction}

Video cameras are increasingly deployed in exploration, monitoring and surveillance applications. These cameras produce vast amounts of information, which needs to be condensed into manageable quantities for both storage and human evaluation. While compression can address the former, this does not aid users, who are often faced with the daunting task of analysing lengthy video sequences or large collections of images. Systems that automatically flag interesting images or information and present a summary to an operator are required to remedy this. This is particularly important in visual search and retrieval applications, where end-users desire highly relevant content, with minimal noise. The ability to predict user preferences reliably is crucial to realising this. 

Unfortunately, it can be hard to define the concept of interesting content, as this is typically context dependent. For example, \citep{castano2005}, which investigates the  feasibility of classifying images by scientific value to address bandwidth constraints on a Mars rover, shows that domain experts from different fields value and rank images differently. As a result, numerous approaches have attempted to build models that can identify content of interest to end-users. These often rely on ranking systems leveraging pairwise comparisons obtained as part of a training phase, but this process can be expensive and time-consuming. 

\begin{figure}
\centering
\includegraphics[width=0.48\textwidth]{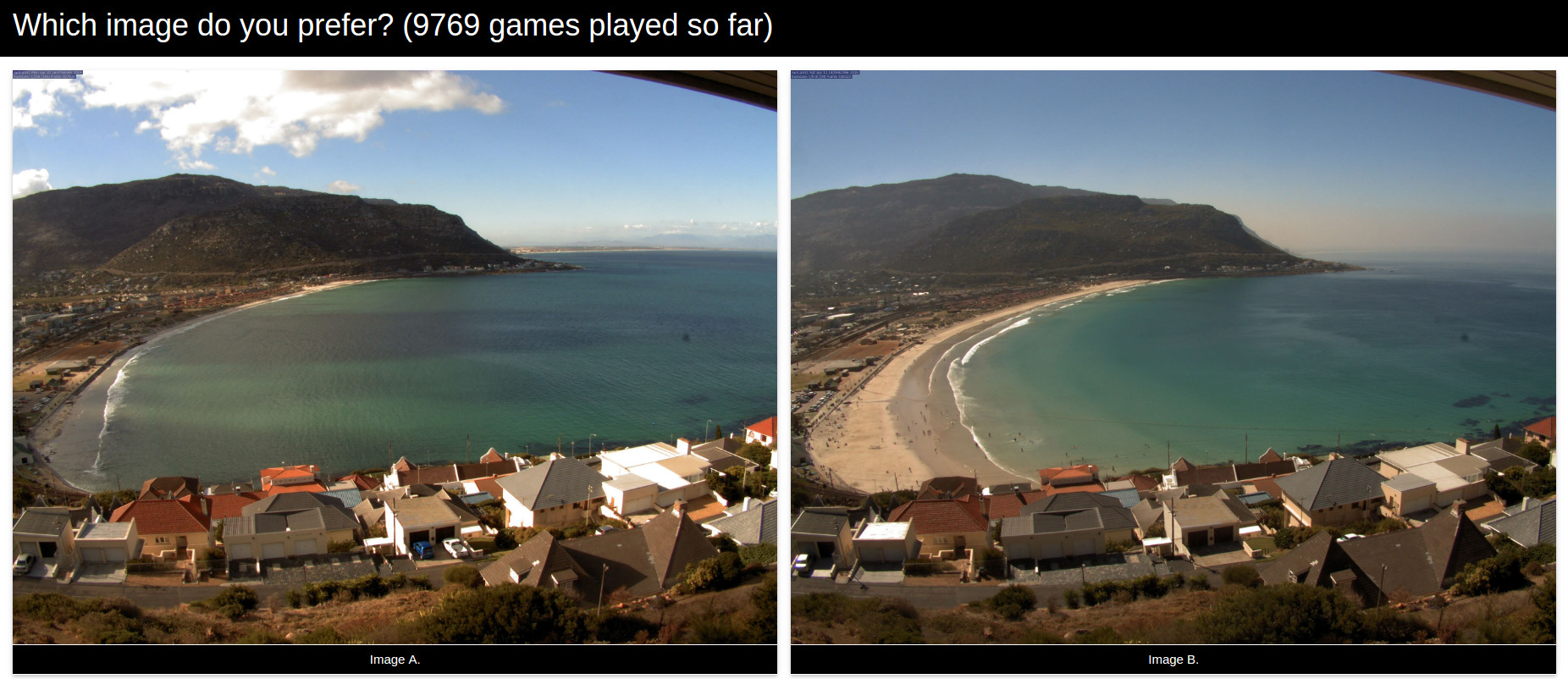}
\caption{A pairwise comparison website is used to source image comparisons suitable for use in a Bayesian ranking system. For the coastal dataset shown here, the right image is preferable, because regions of wet and dry sand are more easily distinguishable than those in the left image.\label{capture_process}}
\end{figure}

More recently, a great deal of work has aimed to develop general models of image interest relying on large general-purpose training databases, in an attempt to avoid retraining models for multiple applications and the need to repeatedly crowd-source training data. However, in this work we argue that domain specific models are still extremely important to end-users. Here, the ability to rapidly train a model suitable for end-user applications with minimal data labelling required is highly desirable. This work introduces a rapid learning approach for domain specific image interest prediction using pairwise image comparisons. Here, pairwise image interest comparisons (Figure \ref{capture_process}) are used to infer image interests using a probabilistic ranking algorithm, and a Gaussian process smoother is then used to improve these estimates by taking into account image similarities using features extracted by a pre-trained convolutional neural network. 

This approach can speed up the learning process significantly, requiring far fewer image comparisons to be labelled to outperform probabilistic benchmark algorithms. In addition, domain-specific models of image interest can be used to produce user-driven storyboards.

The proposed approach targets small-data problems that regularly confront end-users working in specific domains. Here, end-users often need to identify content of interest in small unlabelled datasets, often comprising no more than a few thousand images. These images are often captured at great expense, and the requirements of domain experts and labelling complexity can limit solutions. In this case, pairwise comparison labelling provides a simple, turnkey mechanism of identifying end-user needs, and the design of a problem specific labelling interface is not required.

The primary contributions of this work are as follows:
\begin{itemize}
\item A fully probabilistic image interest estimation scheme is introduced, allowing for image retrieval and ranking with a measure of uncertainty.
\item We show how this measure can be used to determine when sufficient data labelling has been obtained, allowing for sample efficient model training.
\item We show that Gaussian process smoothing in high dimensional image feature space is more effective at image ranking than state of the art neural models when labelling data is limited.
\end{itemize}


\section{Related Work}
\label{related}
As mentioned earlier, the concept of image interest can be rather subjective. This difficulty in defining image interest has led to a wide range of work being conducted in multiple areas seeking to address this topic. We briefly discuss these below, with reference to related work in novelty detection, video storyboarding, image interest and image memorability.    

A common definition of interest relates to novelty, with interest determined by the frequency of occurrence of an event or observation. Novelty detection is often framed as an outlier detection problem. For example, dynamic time warping has been used to align image feature sequences for a life-logging application, with the alignment quality determining novelty \citep{aghazadeh2011}. Here, the authors leverage the fact that people typically experience day-to-day repetition, and assume that areas of mismatch or disagreement with typical daily activity should be flagged as novel. If prior information about the environments or observations to be encountered is available, domain-based approaches to novelty detection are particularly effective. Here, classifiers are trained to recognise expected samples, with any misclassification flagged as novel. For example person, car and groups of person classifiers are trained for a surveillance application by \citet{dieh2002}, with classification failures listed as novel. Terrain classification using support vector machines is applied by \citet{brooks2009}, with negative training data in the form of unlabelled images used to model novelty. 

In contrast to novelty-based image recognition, storyboarding aims to summarise lengthy video sequences using a reduced set of images likely to interest an end-user. This is particularly useful for search and retrieval applications, where users are unwilling to watch a full video in order to evaluate its content. An overview of video storyboarding approaches is provided by \citet{bolanos2016}.

Most storyboarding approaches operate by first segmenting sequences into shots or sub-sequences, and then selecting a representative image for each shot. For example, \citet{Ngo05} use a graph-based clustering approach to segment video into static, panoramic, zoom, motion and in-deterministic shots. An attention model trained on a number of low level features is then used to rank the frames in each shot. This approach provided good performance when the informativeness and enjoyability of the keyframes it produced were evaluated by users. Shots are also used by \citet{srinivasan1999}, with these segmented by detecting changes in image colour histograms. The authors note that scrolling through images is still tedious, so aggregate keyframes selected from shots to form a new video summary of the type typically available for preview in online video repositories.  

MPEG-7 image features have been used in conjunction with image intensity histograms to rank the relevance of images relative to other frames \citep{wolf2005}. Video sequence transitions are detected in \citep{Macer96} by tracking image changes, and selecting keyframes most similar to the average of all frames in shots. Shots selected by detecting video frame transition effects may not be well described using a single key-frame, and a statistical run test is used by \citet{mohanta2013} to segment shots into sub-shots before key-frame selection.

Objects are tracked in image sequences by \citet{guleryuz01}, with images ranked by the length of time objects remain present. A representative frame is selected by finding the frame in each tracked sub-sequence for which the largest number of tracked pixels are present. A people-centric storyboarding approach is taken by \citet{Vonikakis17}, with crowd-sourcing used to identify user preferences when composing slide shows, focusing on facial features and image quality. 

Video storyboarding is of particular interest in life-logging applications, where large amounts of data need to be summarised. Here, egocentric cameras are used to record the daily activities of their wearers. Image sequences of this type often have low temporal consistency, as images are not saved constantly due to storage constraints, so change-based shot segmentation approaches tend to fail. An attempt to remedy this is made by \citet{bolanos2014}, who use an energy minimisation segmentation approach on low level image features to classify images as static, moving camera or in transit. In later work, \citet{mestre2015} use a pre-trained convolutional neural network to identify image features for use in event segmentation for egocentric photo streams.

The storyboarding approaches discussed thus far do not necessarily produce keyframes that are likely to be of interest to humans. In an attempt to remedy this, personalised video summaries are produced by \citet{varini2015} by incorporating a prior on the type of information of interest. Here, a natural language request for images is used to retrieve images in a similar category. Gaze fixation clustering was used by \citet{damen2014multi} to discover areas that are likely to be interesting to humans. Instead of detecting keyframes using novelty, high quality images are found by \citet{xiong2014}. Here, a generative model of `snaps' is trained using an online database of images, under the assumption that most images in online databases are photographs intentionally taken by users and have good composition. Storyboards are formed by segmenting events temporally and selecting keyframes that agree most with this `snap' prior. This approach is particularly effective and has been used for an exploring mobile robot \citep{xiong2014}.

The subjective and contextual nature of image interest makes it hard to design a bottom up interest detection algorithm. Instead, a far more sensible approach makes use of operator supervision to learn about interest. Relative image comparisons are an intuitive way to infer user preference \citep{Hacker2009}, and frequently used for image ranking because they can provide more stable and useful rankings than individual image-based scoring systems \citep{kiapour2014hipster}. 

Pairwise ranking systems are particularly popular across a broad range of problems, and have have been used for optimising visual search \citep{Liu11}, noise reduction in support of highlight detection in video \citep{Kim18} and visual re-ranking in information retrieval \citep{Tian11}. The latter proposes a Bayesian visual re-ranking approach, which re-orders search results using a posterior distribution combining noisy image search results obtained using text queries (a likelihood measure) and an image similarity prior based on block-wise colour moments. Our approach is similar to this, in that we introduce an image similarity prior using a Gaussian process fit over image features extracted using a convolutional neural network, but we combine this with a likelihood inferred from pairwise image comparisons labels returned from end-users instead of queried textual search results. In addition, the use of the Gaussian process prior limits the number of parameters required, as the majority of these are inferred during model training. 

Pairwise ranking is also often used to estimate multimedia quality or predict user preferences. For example, \citep{Ma16} use pairwise ranking to infer image quality from subjective quality score labels, while \citep{Sun17} apply pairwise comparisons to recommend appropriate image filters in social media applications. Here, Amazon Mechanical Turk crowd-sourcing was used to solicit filter preferences from users presented with image pairs in various categories. A convolutional neural network trained to identify image categories was then used to propose suitable image filters, based on the inferred preferences.

A number of effective ranking algorithms have been developed for ranking using pairwise comparisons. Ranking systems such as the Elo chess rating system \citep{elo1978rating} and TrueSkill \citep{herbrich2006trueskill}, a Bayesian ranking scheme extension to Elo, account for relative player skills and performance inconsistency. 

TrueSkill is applied ubiquitously in image ranking systems, providing an effective approach to estimating image interest for a wide range of applications. For example, Hipster wars \citep{kiapour2014hipster} uses TrueSkill to train an image-based style classifier in a fashion application from style judgements, using a part-based model to generate saliency maps that associate clothing items with styles, CollaboRank \citep{janssens2010ranking} uses pairwise comparisons to rank images according to a number of case-based queries (positiveness, perceived threat level, celebrity or film popularity), the Matchin approach \citep{Hacker2009} uses a two player pairwise comparison game to extract a global image `beauty' rank and Streetscore \citep{streetscore} predicts the perceived safety of street scenes using binary answers to the question ``Which place looks safer?". Here, TrueSkill was used to infer street scene safety measures using over 200 000 pairwise image comparisons obtained for approximately 4 000 images. A support vector machine (SVM) was then trained to predict these safety measures using a variety of image features, and then used to build perception maps of city safety in the United States. Unfortunately, this decoupling of SVM interest prediction from the ground truth image interest inference process using TrueSkill means that a highly intensive labelling process is required, with approximately 16 comparisons per image needed to provide interest estimates with high enough levels of certainty for SVM training \citep{streetscore}. This paper shows how this process can be coupled by combining TrueSkill with a Gaussian Process smoother in image feature space, thereby speeding up the labelling process. This coupling is probabilistic and takes interest uncertainty into account so fewer image comparisons are required.  

In contrast to approaches that attempt to infer interest scores from pairwise comparisons, a number of techniques learn to rank directly using these comparisons. These approaches are typically formulated as optimisation problems. For example, \citet{Ma16} learn a linear image feature projection that minimises a binary comparison objective based on image quality, while ranking SVMs \citep{joachims2002optimizing} learn a projection by maximising a Kendall $\tau$ objective (a measure based on the number of concordant and discordant ranked pairs in a list). More recently, this pairwise loss function has been used to train ranking neural networks directly \citep{dubey2016deep,Wang2017RUCAM}, allowing for algorithms that scale to larger datasets, while incorporating the advantages of deep learning. \citet{dubey2016deep} extend Streetscore to consider additional street scene attributes, and capture a significantly larger dataset for experimentation. In order to deal with the challenges of this large dataset, they train a multi-layer neural network to rank image pairs using the ranking SVM loss in combination with an attribute classification loss, and using image features extracted by a pre-trained convolutional neural network. As noted by the authors, coupling the ranking process with image features improves upon traditional two-step processes \citep{dubey2016deep}. However, this approach is not necessarily concerned with the data labelling process, and still assumes that a large representative set of comparisons is already available. In addition, this ranking loss does not account for images that are perceptually similar, for which comparison outcomes may differ when repeated. The probabilistic ranking process described in this paper addresses these challenges.

Pairwise comparisons have also been used to rank abstract paintings according to the emotional responses they elicit \citep{sartori2014affective}, to evaluate the representativeness of images extracted from twitter timelines \citep{wen2014}, and to determine appropriate facial expressions for portraits using images extracted from short video sequences \citep{zhu2014}. Unfortunately, the crowd-sourcing process used to obtain pairwise comparison results can be time consuming and expensive \citep{turkpay} and a large number of comparisons are typically required to infer interests. In an attempt to remedy this, heuristic budget constraints are introduced into a pairwise ranking process by \citet{Cai17}, while \citet{Burke16} proposes a smoothing algorithm that uses the temporal image interest similarity present in video to improve interest estimates with fewer comparisons. The latter relies on a Markovian assumption, and so fails to account for interest similarity that is likely to occur when images are captured in the same place at different times, or if images themselves appear similar. This paper introduces a Gaussian process smoother that addresses this limitation.

More recently, there have been attempts to train more general models of image interest, most notably for the 2016 \citep{shen2016technicolor} and 2017 Predicting Media Interestingness MediaEval challenges \citep{interestingness2017}. For the 2017 task, interestingness is defined within the context of extracting frames and film excerpts that would aid a user to make a decision about whether they would be interested in watching a movie. This task is relatively general purpose, as movies cover different topics and genres, but inevitably favours aesthetics and genre or emotional content in the definition of interest. As a result, prediction methods that introduce genre prediction systems and related contextual information tend to perform well on this task. For example, Ben-Ahmed et al. \citep{Ahmed17} use a deep neural network to predict genres from image interests, and a SVM to predict genres from audio features. The genre logits obtained from these models are then used as a multimedia representation, and a final SVM is trained using these to predict a binary image interest value. Berson et al. \citep{Berson17} use a broad range of information (image features, image captioning representations, audio features, and representations extracted from textual meta-data)  within a large multimodal neural network framework to predict a binary image interest value, noting that the inclusion of contextual information like image captions and textual meta-data can lead to over-fitting on individual image interest prediction tasks, but improved performance on video interest prediction. 

The Predicting Media Interestingness challenge was adapted to become a memorability prediction challenge in 2018 \citep{MediaEval18}. Memorability is closely related to image interest, and typically measured using an experimental approach where users are shown a sequence of images, with some repeated, and asked to recall which images they have seen previously. \citet{Khosla} carried out a comprehensive study of memorability and made an extremely large database of memorability scores and associated images available. Here, image memorability was shown to relate to image popularity and emotional content, but not necessarily to aesthetics. 

While an effective measure of image interest, memorability may be unsuited for domain-specific small to medium scale computer vision problems, as the labelling burden on end-users can be excessive. This work seeks to highlight the subjective nature of image interest through a number of domain-specific cases and to emphasise that for many use cases, domain-specific models of interest are needed. This typically requires an intensive labelling process, but this work shows that a Gaussian process smoother combined with a Bayesian ranking system can infer image interest scores in a stable and efficient manner, providing information about interest prediction certainty, thereby facilitating more rapid deployment of models.

\section{Image interest estimation}
\label{interest}

Our goal is to use pairwise image comparisons to train a model that can predict image interest. This model can then be used for image storyboarding. Initially, a baseline Bayesian ranking scheme is used to estimate image interest scores. This is combined with a Gaussian process smoother that improves estimates by incorporating image similarity information from convolutional neural network image features. We compare this probabilistic approach with a deep learning approach using a pairwise loss function.

\subsection{Probabilistic image ranking}
\label{ranking}

This work uses the TrueSkill Bayesian ranking scheme \citep{herbrich2006trueskill} to compute image interest scores. TrueSkill is a probabilistic ranking system that assumes players in a game have respective skills, $w_1$ and $w_2$, and that game outcomes can be predicted by the performance difference between skills, subject to Gaussian noise effects.

For image pairs,
\begin{equation}
t \sim \mathcal{N}(s,1) 
\end{equation}
models the interest difference between two images, with $s = w_1 - w_2$ the interest difference and the standard normal distribution accounting for potential labelling errors \citep{Burke16}. Comparison outcomes are given by $y = \text{sign}(t)$, with a positive $y$ indicating a win for image 1, and a negative $y$ indicating a loss.

Interest estimation under this model can be treated as a Bayesian inference problem, with the posterior over skills described by
\begin{equation}
p(w_1,w_2|y) = \frac{p(w_1)p(w_2)p(y|w_1,w_2)}{\int\int p(w_1)p(w_2)p(y|w_1,w_2)\text{d}w_1\text{d}w_2} ,
\end{equation}
where $p(w_i) = \mathcal{N}(\mu_i,\sigma^2_i)$ is a Gaussian prior over image interests and 
\begin{equation}
p(y|w_1,w_2) = \int\int p(y|t)p(t|s)p(s|w_1,w_2)\text{d}s\text{d}t
\end{equation}
the likelihood of a game outcome given interests. The model above is easily extended to multiple images, $\mathbf{w}$, by chaining comparisons, $\mathbf{y}$, together in a large graph, producing the posterior $p(\mathbf{w}|\mathbf{y})$. This posterior is intractable, but can be estimated numerically and approximated by a Gaussian \citep{minka2001family}
\begin{equation}
p(\mathbf{w}|\mathbf{y}) \sim \mathcal{N}(\mathbf{w}_m,\mathbf{\Sigma}_n), \label{eq_post}
\end{equation}
with mean $\mathbf{w}_m$ and variance $\mathbf{\Sigma}_n$.

\subsection{Temporal TrueSkill}

The interests inferred using TrueSkill are only updated for those images involved in pairwise comparisons. As a result, a large number of comparisons could be required to infer interest values to an acceptable level of certainty when image datasets are large. However, where image interests are required for image sequences or video datasets, a simple posterior smoothing process \citep{Burke16}, hereafter referred to as temporal TrueSkill (TTS), can be used to improve the TrueSkill estimates.

Here, image interests in a video sequence are assumed to follow a random walk motion model $p(x_k|x_{k-1})$, and image distributions inferred using TrueSkill used as measurement models for the $k$-th image in a sequence of $K$ images, $p(w_k|x_k)$, within a standard Rauch-Tun-Striebel smoother \citep{rauch1965}, to provide a posterior distribution over image interests, conditioned on a sequence of TrueSkill estimates, $p(x_k|w_{1:K})$,
\begin{align}
p(x_k|w_{1:k-1}) &= \int p(x_k|x_{k-1})p(x_{k-1}|w_{1:k-1}) \text{d}x_{k-1}\nonumber\\
p(x_k|w_{1:k}) &= \frac{p(w_k|x_k)p(x_{k-1}|w_{1:k-1})}{p(w_k|w_{k-1})}\nonumber\\
p(x_k|w_{1:K}) &= \int\frac{p(x_{k+1}|x_k)p(x_k|w_{1:k})}{p(x_{k+1}|w_{1:k})}\times\nonumber\\& p(x_{k+1}|w_{1:K})\text{d}x_{k+1}.
\end{align}
  
Temporal TrueSkill is computationally inexpensive, but fails to account for similarities with images themselves. The Gaussian process (GP) interest refinement proposed here addresses this limitation.

\subsection{Gaussian process interest refinement}
\label{GPSmooth}

As an alternative to the smoothing algorithm used for TTS, this work refines image interest estimates obtained using TrueSkill using a Gaussian process smoother operating in image feature space. A GP is a collection of random variables, where any finite number have a joint Gaussian distribution \citep{rasmussen2006gaussian}. Gaussian processes,
\begin{equation}
f(\mathbf{x}) \sim \mathcal{GP}(m(\mathbf{x}),k(\mathbf{x},\mathbf{x'})),
\end{equation}
are specified by the mean function $m(\mathbf{x})$ and the covariance function $k(\mathbf{x},\mathbf{x'})$ of a real process $f(\mathbf{x})$,
\begin{align}
m(\mathbf{x}) &= \mathbb{E}[f(\mathbf{x})]\\
k(\mathbf{x},\mathbf{x'}) &= \mathbb{E}\left[\left(f(\mathbf{x}) - m(\mathbf{x})\right)\left(f(\mathbf{x'}) - m(\mathbf{x'})\right)\right].
\end{align}

For the image interest application, the domain $\mathbf{x}$ is over a set of image attributes or features associated with an image, while $f$ is the process that gives rise to image interest. $\mathbf{x}'$ denotes the features or attributes associated with captured image interest random variables $\mathbf{w} = [w_1 \dots w_N]$, where $N$ denotes the number of images. The mean function $m(\mathbf{x})$ is assumed to be zero in this work. 

Under this process, a likelihood for image interests, $\mathbf{w}$, can be formed, 
\begin{equation}
p(\mathbf{w}|\mathbf{x},f) \sim \mathcal{N} \left(f(\mathbf{x}), \mathbf{\Sigma}(\mathbf{x})\right).
\end{equation}
Using this likelihood in conjunction with a GP prior,
\begin{equation}
p(f) \sim \mathcal{GP}(\mathbf{0},k(\mathbf{x},\mathbf{x'})),
\end{equation}
and taking advantage of the marginalisation properties of Gaussian processes, leads to a Gaussian process posterior \citep{rasmussen2006gaussian},
\begin{equation}
p(f|\mathbf{x},\mathbf{w}) \sim \mathcal{GP}(m_\text{p},k_\text{p}), \label{GPpost}
\end{equation}
where
\begin{align}
m_\text{p} &= \mathbf{T}(\mathbf{X},\mathbf{X'})\mathbf{w}_m,\\
k_\text{p} &= K(\mathbf{X},\mathbf{X'}) - \mathbf{T}(\mathbf{X},\mathbf{X'})K(\mathbf{X},\mathbf{X'}),
\end{align}
and
\begin{equation}
\mathbf{T}(\mathbf{X},\mathbf{X'}) = K(\mathbf{X},\mathbf{X'})[K(\mathbf{X'},\mathbf{X'}) + \mathbf{\Sigma}(\mathbf{X'})]^{-1}.
\end{equation}
Assuming $N$ training images with features $\mathbf{X}$, and $N'$ query images with features $\mathbf{X'}$, $K(\mathbf{X},\mathbf{X'})$ denotes the covariance matrix formed by evaluating $k(\mathbf{x},\mathbf{x'})$ for all pairs of training and test features. $\mathbf{\Sigma}(\mathbf{X'}) = \mathbf{\Sigma}_n$ is a diagonal matrix with diagonals corresponding to the variance in estimated image interests $\mathbf{w}_m$, obtained from the TrueSkill posterior in (\ref{eq_post}). Equation (\ref{GPpost}) can be used for interest prediction by evaluating the GP posterior for a set of images with features $\mathbf{X}^*$,
\begin{align}
p(\mathbf{w}^*|\mathbf{X}^*,\mathbf{X'},\mathbf{w}) &\sim  \\\mathcal{N} (\mathbf{T}(\mathbf{X}^*,\mathbf{X'})\mathbf{w}_m,& K(\mathbf{X}^*,\mathbf{X}^*) \mathbf{T}(\mathbf{X}^*,\mathbf{X'})K(\mathbf{X}^*,\mathbf{X'})). \nonumber\label{heterogp}
\end{align} 

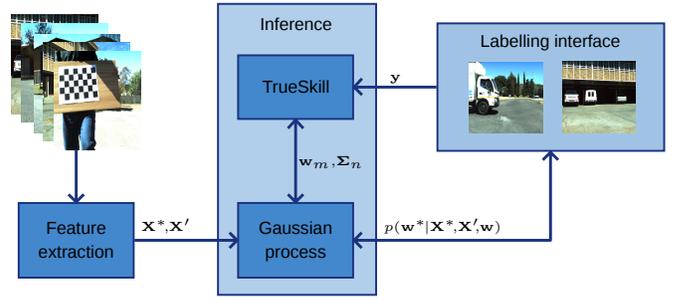
\begin{figure}
\centering
\begin{overpic}[width=0.48\textwidth]{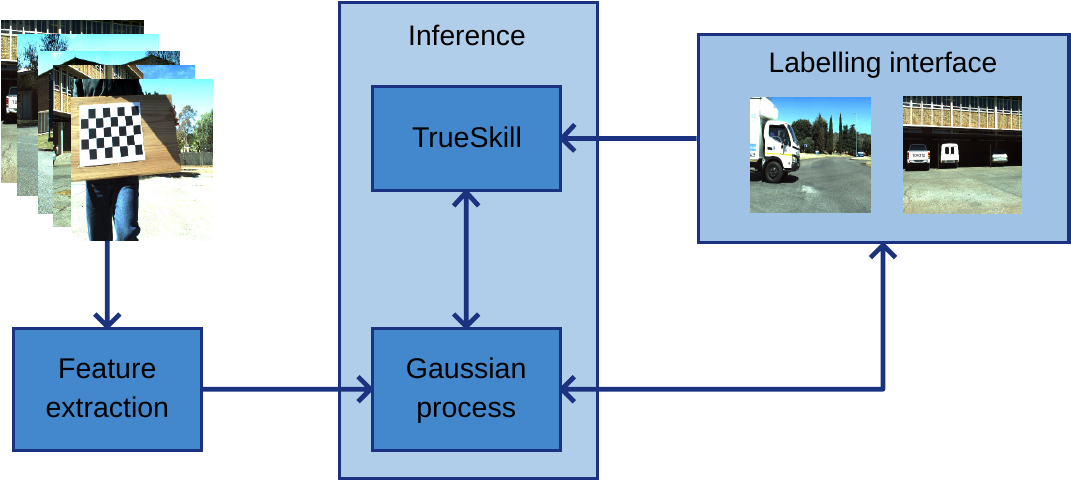} 
\put(20,10){\tiny{$\mathbf{X}^*\hspace{-0.5mm},\hspace{-0.5mm}\mathbf{X'}$}}
\put(57,10){\tiny{$p(\mathbf{w}^*\hspace{-0.3mm}|\mathbf{X}^*\hspace{-0.5mm},\hspace{-0.5mm}\mathbf{X'}\hspace{-0.7mm},\hspace{-0.5mm}\mathbf{w})$}}
\put(44,20){\tiny{$\mathbf{w}_m,\hspace{-0.5mm}\mathbf{\Sigma}_n$}}
\put(58,33){\tiny{$\mathbf{y}$}}
\end{overpic}
\caption{The GP-TS image interest prediction process is depicted above. Input images are fed into a deep convolutional neural network, producing a $d$-dimensional feature vector. This feature vector is then fed into a Gaussian process that is trained using image features and corresponding TrueSkill image interest estimates, inferred using pairwise comparison labels.\label{GP_CNN_Alg}}
\end{figure}

A wide variety of covariance functions can be used, but for this work we apply a radial basis function kernel to ensure smooth interests over image feature space,
\begin{equation}
k(\mathbf{x},\mathbf{x'}) = \exp{\left(-\frac{D(\mathbf{x},\mathbf{x}')}{2l^2}\right)}.
\end{equation}
Here, $l$ is a length scale hyperparameter used to control the level of similarity at which image attributes affect one another, and $D$ is a distance measure appropriate to the image attributes selected for smoothing. The image attributes considered here comprise $d$-dimensional image features extracted using a pre-trained convolutional neural network \citep{szegedy2015}, while the cosine distance,
\begin{equation}
D(\mathbf{x},\mathbf{x}') = 1 - \frac{\mathbf{x} \cdot \mathbf{x}'}{\lVert \mathbf{x} \rVert \lVert\mathbf{x}'\rVert},
\end{equation}
is used as the distance measure. Figure \ref{GP_CNN_Alg} illustrates the image interest inference and smoothing approach described above, referred to as GP-TS hereafter.

Gaussian processes are memory intensive, $\mathcal{O}(N^3)$, so are often considered unsuitable for large image datasets. However, given that our goal is to learn about image interest for the small-data regime where limited numbers of images and labels are required, this is typically not problematic. For larger datasets, sparse Gaussian processes \citep{herbrich2003fast} or Bayesian committee machines \citep{tresp2000bayesian} reduce this complexity significantly.

\subsection{GP-TS Inference}

We consider a number of approaches to perform probabilistic inference under the GP-TS model. The first decouples inference using the Gaussian process and Trueskill, with inference performed separately for each component. Here, inferred image interest levels are initially estimated using expectation propagation \citep{minka2001family} under the Trueskill model. Expectation propagation approximates factors in the model using Gaussian distributions fit through moment matching, which allows for efficient inference by message passing. This produces the approximate posterior in (\ref{eq_post}), with mean interest estimates and uncertainties for each image in the set conditioned on image comparison outcomes. This distribution over image interests can then be used to perform inference under a Heteroscedastic Gaussian process model \citep{le2005heteroscedastic}, with the length scale parameter $l$ inferred using maximum a-posteriori estimation. 

As an alternative, inference under the GP-TS model can be treated in a fully Bayesian manner with appropriate priors over parameters. In this case, we construct the GP-TS generative model as follows, with parameter definitions unchanged from previous sections:
\begin{align}
l &\sim \text{Half Cauchy} (\beta=0.5)\nonumber\\
\mathbf{\Sigma}_n &\sim \text{Half Cauchy} (\beta=1)\nonumber\\
\mathbf{f}(\mathbf{x}) &\sim \mathcal{GP}(\mathbf{0},k(\mathbf{x},\mathbf{x'}))\nonumber\\
\mathbf{w}^* &\sim \mathcal{N}(\mathbf{f},\mathbf{\Sigma}_n)\nonumber\\
p &= \text{Sigmoid} (w^*_i - w^*_j)\nonumber\\
y &= \text{Bernoulli} (p).
\end{align}

Here, length scale $l$ and interest uncertainty $\mathbf{\Sigma}_n$ are modelled using half Cauchy priors. The zero-mean Gaussian process prior over features extracted from images using a pre-trained convolutional neural network is used to model image interest. The marginal likelihood of this prior, which incorporates labelling inconsistency noise, provides a predictive distribution for image interests given image features. Comparison outcomes are modelled as a Bernoulli trial given a probability formed by passing the difference in interests ($w^*_i$) and ($w^*_j$), between the image pairs through a sigmoid function. This model allows for variational Bayesian inference strategies such as automatic differentiation variational inference \citep{kucukelbir2017automatic} to be applied. Like expectation propagation, variational inference approximates distributions using a family of simpler distributions, framing inference as a task of minimising the Kullback-Liebler divergence of samples from the posterior (training data) from the simpler target distributions. This approach allows for efficient parallel batch estimation, leveraging many advances in gradient-based optimisation for deep learning. In this work, we use the PyMC3 probabilistic programming library \citep{salvatier2016probabilistic} for inference. 

Inference in the fully Bayesian setting can be expensive, so we also consider the use of Gaussian process approximations such as sparse Gaussian processes \citep{herbrich2003fast}, which rely on factorisation to reduce the computational complexity of GP's to $\mathcal{O}(NM^2)$. Here, $M$ is a parameter controlling the number of input features to use for estimating the Gaussian process kernel.

\subsection{Pairwise loss ranking}

A deep learning approach, trained directly using pairwise comparisons to minimise a pairwise loss function \citep{dubey2016deep,Wang2017RUCAM} can be used as an alternative to the probabilistic approaches described above. Here, image features are first extracted from each image in a comparison pair using a pre-trained convolutional neural network. These features are then fed into two weight-tied multi-layer fully connected neural networks (typically 2-3 layers using ReLU activation functions) producing scalar outputs $y$ and $x$, and trained to minimise the loss,
\begin{equation}
\text{PWL} = \sum_{i=1}^n \text{ReLU}(y-x),
\end{equation} 
 using stochastic batch gradient descent. This loss is equivalent to a ranking SVM loss \citep{joachims2002optimizing}, but has been simplified here by assuming that the comparison winner is always input to the network producing $y$. This approach is referred to as FC-PWL hereafter.

\section{Storyboarding}

The image interest estimates obtained using pairwise ranking systems are easily used for storyboarding. This is a simple matter of selecting $N_s$ images corresponding to the top mean image scores, requiring that these are at least $d_s$ images apart for sequential datasets. Here, both $d_s$ and $N_s$ are left as user defined input parameters, to allow for customised and controllable storyboarding. Giving a user the ability to adjusting these parameters and display relevant results within an exploration tool is a particularly effective means of exploring image datasets. 

A similar approach can be taken to produce image memorability-based storyboards. In this work, we compare GP-TS storyboards with those produced using a pre-trained image memorability predictor, MemNet \citep{Khosla}. MemNet is a deep convolutional neural network trained using 60 000 images sampled from a number of image collections (both scene and object-centric) and corresponding memorability scores, captured using an intensive labelling process.

As an alternative to storyboarding using image interest or memorability, clustering approaches to storyboarding attempt to summarise image datasets by finding a representative set of images. In this work, we also compare GP-TS and MemNet storyboarding with a recent clustering approach \citep{mestre2015}. Here, hierarchical agglomerative clustering \citep{ward1963} is applied to the same pre-trained convolutional neural network image features used by GP-TS. After grouping images into $N_s$ clusters, a representative image is selected for each cluster by finding the image with a feature vector closest to the mean image feature vector for each cluster. This clustering approach to storyboarding is termed HAC hereafter.

\section{Datasets}

The proposed approach to turnkey image interest estimation and storyboarding was investigated using five distinct datasets. Each of these is briefly described below.

\subsection{OASIS}

The first dataset used for testing is a small publically available medical imaging dataset of 416 averaged and co-registered T1-weighted cross-sectional magnetic resonance imaging scans of patients with varying levels of dementia \citep{marcus2007}. The scans are normalised and accompanied by meta-data that includes normalised brain volume measurements. Pairwise comparison results were simulated by generating 15 000 comparison outcomes using the normalised brain volume measurements. Here, we assume that brain volume reductions correlate with those images of patients depicting reduced brain matter, and that a domain expert would consider images with reduced brain matter of importance. The 15 000 comparison results, $\mathbf{G}_\text{baseline}$, were split into test, $\mathbf{G}_\text{test}$, and training, $\mathbf{G}_\text{train}$, sets, comprising 5 000 and 10 000 comparisons respectively.

\subsection{Violence}

The second dataset used for testing is a publically available dataset of over 10 000 protest images \citep{won2017protest}, with accompanying measures of the perceived violence depicted therein. As before, pairwise comparison results were simulated by generating 15 000 comparison outcomes using these perceived violence scores. Here, it was assumed that an end-user would be interested in identifying scenes depicting violence. Unlike the dataset above, the perceived violence dataset is already divided into test (2 342 images) and training (9 316 images) sets. In order to align with this division, we split the 15 000 comparison results obtained from the training set,  $\mathbf{G}_\text{baseline}$, into 5 000 test examples, $\mathbf{G}_\text{test}$, and 10 000 training examples, $\mathbf{G}_\text{train}$, but also generated an additional test set, $\mathbf{G}^2_\text{test}$, of 10 000 comparisons using images sampled at random from the perceived violence test images, $\mathbf{G}_\text{baseline\_test}$. Results are reported for both of these test sets.
 
\subsection{CSIR}

The third dataset comprises 4 000 outdoor images captured by an autonomous rover containing a sequence of images captured in an uncontrolled outdoor environment. Here, 15 000 baseline pairwise image comparison results, $\mathbf{G}_\text{baseline}$,  were obtained using a labeling interface (Figure \ref{capture_process}) that presented randomly selected pairs of images to a single robot operator and asked which image was more useful to them. In general, the robot operator (wary of potential collisions) favoured images that contained cars or pedestrians. As before, the 15 000 baseline image comparisons were split into test, $\mathbf{G}_\text{test}$, and training, $\mathbf{G}_\text{train}$, sets, comprising 5 000 and 10 000 comparisons respectively.

\subsection{Coastcam}

The fourth dataset consists of almost 2 000 outdoor images of the Fishhoek coastline in South Africa, captured from a static camera \citep{Burke16}. Here, 10 000 baseline pairwise image comparison results, $\mathbf{G}_\text{baseline}$,  were obtained by presenting randomly selected pairs of images to a single domain expert and asking which image was more important (Figure \ref{capture_process}). The domain expert favoured images that showed images where wet and dry sand regions were clearly identifiable. As before, the baseline image comparisons were split into test, $\mathbf{G}_\text{test}$, and training, $\mathbf{G}_\text{train}$, sets, comprising 3 300 and 6 700 comparisons respectively.

\subsection{Place Pulse 2.0}

The final dataset used for testing comprises 110 988 Google Streetview images taken from 56 cities \citep{dubey2016deep}. Here, over 1 million baseline pairwise image comparisons were captured and made publically available for six perceptual attributes: safe, lively, boring, wealthy, depressing and beautiful. In this work, only the safety attribute is considered, with 323 392 comparisons. These baseline image comparisons $\mathbf{G}_\text{baseline}$ were split into test, $\mathbf{G}_\text{test}$, and training, $\mathbf{G}_\text{train}$, sets, comprising 106 720 and 216 672 comparisons respectively. This dataset is used to test the scalability of the proposed approach in ensemble form.

\section{Experimental results}
\label{GPresults}

\begin{figure*}
\centering
\includegraphics[width=\textwidth]{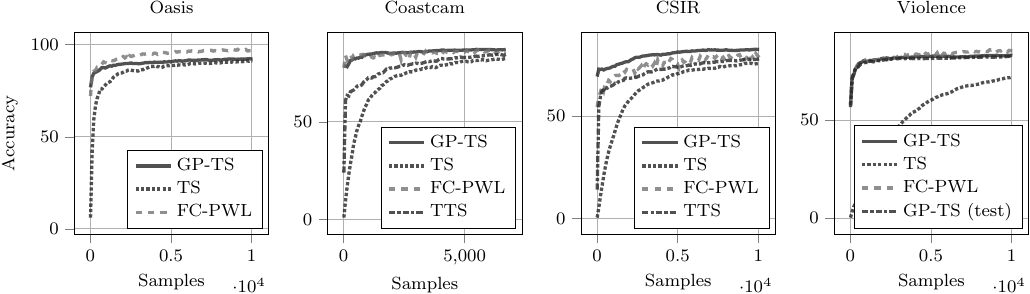}
\caption{Traces of the image comparison prediction accuracy as the number of samples used for model training is increased highlight the performance of GP-TS. Note that temporal TrueSkill (TTS) was only used on the video datasets, as this approach requires sequential data. As the perceived violence dataset is already divided into test and train sets, we report the game prediction accuracy using test sets,  $\mathbf{G}_\text{test}$ and $\mathbf{G}^2_\text{test}$, extracted from both training, $\mathbf{G}_\text{baseline}$, and test sets, $\mathbf{G}_\text{baeline\_test}$, and a model trained on an increasing number of pairwise labels extracted from the training set,  $\mathbf{G}_\text{train}$. \label{pred_acc}}
\end{figure*}

\begin{figure}
\centering
\includegraphics[width=0.45\textwidth]{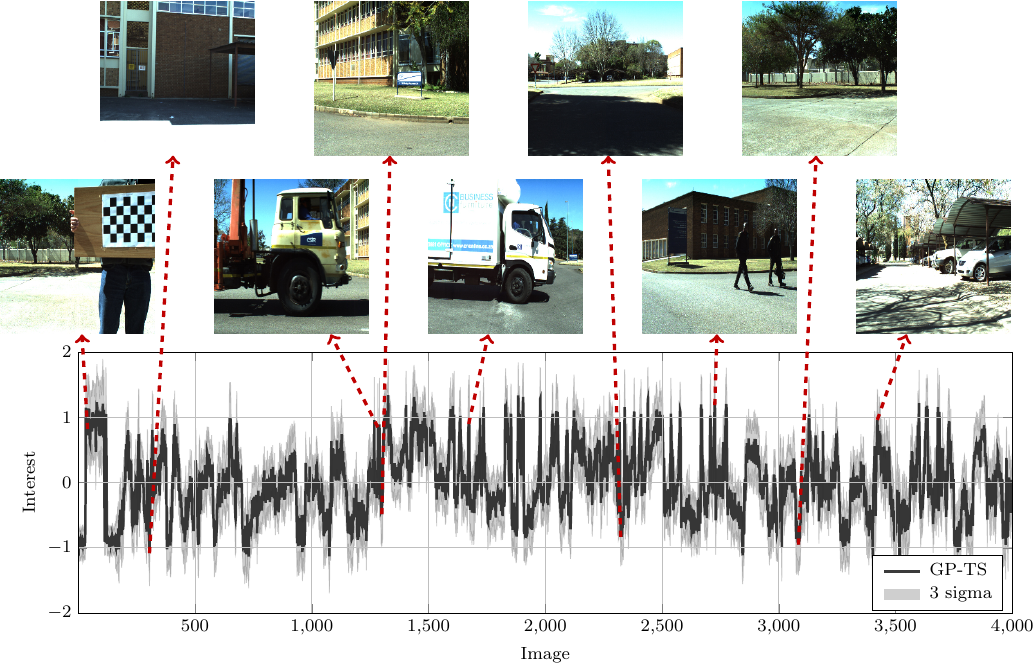}
\caption{Image samples with higher interest scores tend to contain vehicles or pedestrians (in line with the operator's preference), while image samples with lower interest scores are generally empty road scenes or images of buildings.\label{interest_pics}}
\end{figure}

\begin{figure*}
\centering
\includegraphics[width=\textwidth]{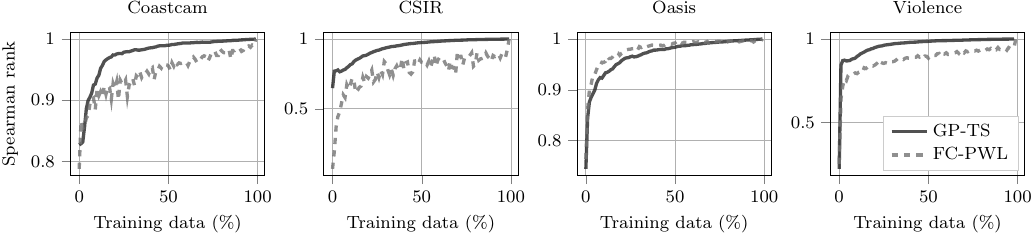}
\caption{Spearman rank correlations between final rankings obtained using all training data and those trained with limited labelling show that GP-TS tends to converge to the true rank faster than a FC-PWL, indicating that it is more sample efficient. \label{spearmanrank}}
\end{figure*}

\subsection{GP-TS inference strategies}

A number of inference strategies for GP-TS were evaluated using the CSIR dataset. These include decoupled heteroscedastic GP-TS inference (DH-GP-TS), decoupled heteroscedastic GP-TS inference using sparse GP's (DH-SGP-TS), automatic differentiation variational inference under the fully Bayesian GP-TS model (ADVI-GP-TS) and automatic differentiation variational inference under the fully Bayesian GP-TS model using sparse GP's (ADVI-SGP-TS). Inception V3 bottleneck features were used for GP covariance function evaluations. 

Table \ref{inference_strategies} shows the comparison prediction accuracy obtained using each of these approaches, when all available comparison outcomes were used for inference, and trained models used to predict comparison outcomes in the test set. The number of iterations used for inference are denoted by $k$, while $M$ denotes the number of inducing image features used by the sparse Gaussian process. These features are selected by $K$-means clustering the image features in the training set. Prediction accuracy refers to the fraction of game outcomes that were correctly predicted by computing the posterior predictive probability of each image winning a comparison game outcome. This probability is thresholded, under the assumption that a game outcome is correct if the predicted probability in favour of the image winning the game is greater than 50 $\%$.

Interestingly, decoupling the inference phases proved far more effective than performing inference under a fully Bayesian model, presumably because the inference task is simplified dramatically through this decoupling, as evidenced by the small number of expectation propagation iterations ($k$) required for inference in this case. The sparse GP approximation produces a moderate performance drop, but with substantial reduction in computational time. In light of these results, all experiments are conducted using DH-GP-TS for the remainder of this paper, which is termed GP-TS for brevity. 
\begin{table}
\centering
\caption{Inference strategy efficacy \label{inference_strategies}}
\begin{tabular}{|l|r|c|c|}
\hline
 & \bf{Parameters} & \bf{Time} & \bf{Acc.}\\
 & & (mm:ss) & (\%) \\
 \hline
 \bf{DH-GP-TS} &k=5 & 1:56 & 80.89\\
 \hline
 \bf{DH-SGP-TS} &k=5, M=100 & 0:47 & 77.43\\
 \hline
 \bf{ADVI-GP-TS}  &k=200 & 45:18 & 74.79\\
 \hline 
\bf{ADVI-SGP-TS} &k=200, M=100 & 1:24 & 73.56\\
 \hline
\end{tabular}
\end{table}

\subsection{Interest prediction}

Four interest detection algorithms were compared: A TrueSkill interest estimate (TS) \citep{herbrich2006trueskill}, a temporally smoothed interest algorithm (TTS) \citep{Burke16}, the proposed GP interest estimation approach, GP-TS, and a deep pairwise ranking approach, FC-PWL. Both GP-TS and FC-PWL use image features extracted using the Inception-V3 convolutional neural network, pre-trained for image classification on the ImageNet database \citep{szegedy2015}. The FC-PWL model uses 3 fully connected layers comprising 2 048, 1 024 and 1 neurons respectively, and was trained for 50 epochs using the Adam optimser with parameters defined as in \citep{kingma2014adam} and a batch size of 256. These parameters were chosen because they produced the most reliable results across all datasets.

Figure \ref{pred_acc} shows traces of the image comparison prediction accuracy for each algorithm, on each of the first four test datasets. Here, an increasing number of comparisons sampled from training sets, $\mathbf{G}_\text{train}$, were used to predict game outcomes for the comparison pairs in $\mathbf{G}_\text{test}$, for each of the four datasets. Note that the results of the proposed approach are also shown for the test set of the violence dataset, $\mathbf{G}^2_\text{test}$, but with models still trained using subsets of the training set,  $\mathbf{G}_\text{train}$. In the case of the non-probabilistic FC-PWL approach, game winners were predicted by selecting the image producing the largest logit predicted by the neural network pairs.

Figure \ref{interest_pics} shows the posterior predictions for GP-TS when all 15 000 comparisons are used for interest estimation on the CSIR dataset, along with a selection of images corresponding to various interest levels. Images with higher interest scores contain objects of interest (pedestrians or vehicles), while images with lower image interest scores are more likely to be of empty road scenes.

It is clear that GP-TS outperforms the interest estimation of TTS and TS. Smoothing in image feature space requires significantly fewer training comparisons to outperform the baseline probabilistic interest prediction algorithms. TTS results are only provided for sequential image datasets, as this approach requires video or image sequences. FC-PWL performs similarly to GP-TS, outperforming the latter on the simpler OASIS dataset, but under-performing on the CSIR dataset. It should be noted that FC-PWL needed to be hand tuned to find parameters that worked across each dataset, relying on neural network designer skills and experience to do so. In contrast, the GP-TS approach requires no design expertise, as all parameters are inferred automatically.

More importantly, the GP-TS approach is more sample efficient, and produces better ranking estimates with limited labelling data. This is visible when the Spearman rank correlation is measured between the image interests inferred using only a portion of the training data, and those inferred using all available data (Figure \ref{spearmanrank}). This is true for all but the Oasis dataset, which is simple enough to rank using relatively few image comparisons.
\begin{figure}
\centering
\includegraphics[width=0.5\textwidth]{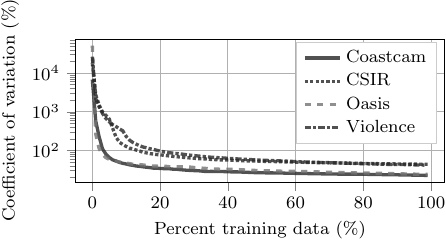}
\caption{Coefficient of variation curves as a function of training data can be used to evaluate model performance and labelling data requirements. \label{pred_uncertainty}}
\end{figure}

\begin{figure}
\centering
\includegraphics[width=0.5\textwidth]{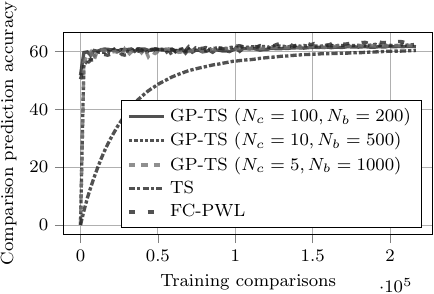}
\caption{Traces of the image comparison prediction accuracy as the number of samples used for model training is increased show that an ensemble of GP-TSs exhibits similar convergence results to those obtained using individual regressors. \label{pred_acc_ensemble}}
\end{figure}

\subsection{Uncertainty analysis}

The combination of the Gaussian process with TrueSkill means that GP-TS is a probabilistic model and image interest predictions are paired with a variance measure. This measure captures the uncertainty in an interest prediction, but also uncertainty due to inconsistent labelling, which may occur due to labelling error, or simply because images compared have similar interest values. These probabilistic estimates are particularly valuable, as they can be used to propose comparisons to present within an active labelling framework, or to select interesting content to show to users while taking into account the potential uncertainty therein. Figure \ref{pred_uncertainty} shows the average coefficients of variation (the average ratio of the predicted standard deviation to the absolute value of the predicted mean interest) as a function of the number of pairwise comparisons used for inference using each of the test datasets. As expected, the predictions become more certain (less volatile) with additional comparisons. Convergence to a stable estimate is obtained after relatively few comparisons. The accompanying video shows how uncertainty and interest changes during the training process.

The ability to estimate the uncertainty in inferred image interests is particularly valuable, as it can be used as a convergence measure to decide when enough comparisons have been captured during a dataset labelling process. Current state of the art methods such as FC-PWL, which only provide point-estimate predictions, require that a large test set be captured in order to test model accuracy and evaluate algorithm performance so as to determine how much labelling data is required to train a reliable model. Further, there are no guarantees regarding the certainty in individual image interest predictions using these approaches and no exisiting mechanisms for determining when sufficient labelling has occured. 

\subsection{Scaling to large datasets}

As mentioned previously, Gaussian processes are often deemed unsuitable for large datasets as they are memory intensive. However, ensemble approaches can be used to remedy this. Figure \ref{pred_acc_ensemble} shows the results obtained when an ensemble of GP-TSs is used to predict the perceived safety of a street scene using training data sampled from the Place Pulse 2.0 dataset \citep{dubey2016deep}. Experimental results provided follow the same procedures as before, but here $N_e$ Gaussian processes were trained to predict TrueSkill interests using batches of $N_b$ images sampled from the dataset. It is clear that the ensemble exhibits similar convergence results to those seen previously, and is relatively robust to parameter choices.

Table \ref{tab_auc} shows the percentage area under the curve (relative to the maximum possible area) for each method on the various datasets of interest, and provides ablation results when the pre-trained features used as inputs to GP-TS are varied. Here, Inception-V3 \citep{szegedy2015}, Resnet50 \citep{he2016deep}, VGG16 \citep{simonyan2014very} and Histogram of Oriented Gradient (HoG) \citep{dalal2005histograms} features are used for testing. 

GP-TS and FC-PWL perform similarly with less training data, but, as expected, FC-PWL performance improves when substantially more data is available. Ablation results show that pre-trained convolutional network and HoG features used by GP-TS produce generally similar results, although HoG performance drops for more challenging datasets. Due to computational limitations, experiments on Place Pulse 2.0 were only conducted using Inception-V3 features.
\begin{table*}
\centering
\caption{\% Area under curve (prediction accuracy vs training data) \label{tab_auc}}
\begin{tabular}{|l|c|c|c|c|c|}
\hline
 & \bf{OASIS} & \bf{Violence} & \bf{CSIR} & \bf{Coastcam} & \bf{Place Pulse 2.0}\\
 \hline
 \bf{GP-TS} & & & & & \\
 \hline
 \hspace{0.2mm} HoG & 89.29 & 69.12 & 72.57 & 84.87 & -\\
 \hline
 \hspace{0.2mm} Inception-V3 & 90.27 & 80.43 & \textbf{79.86} & 85.43 & 60.84\\
 \hline
 \hspace{0.2mm} ResNet50 & 90.85 & 82.44 & 78.81 & \textbf{85.85} & -\\
 \hline
 \hspace{0.2mm} VGG16 & 90.69 & 81.32 & 78.48 & 85.35 & -\\
 \hline
 \bf{TS} & 85.60 & 52.66 & 63.84 & 70.86 & 51.25\\
 \hline
 \bf{TTS} & - & - & 72.22 & 78.08 & -\\
 \hline
 \bf{FC-PWL} & \textbf{94.59} & \textbf{82.83} & 75.20 & 84.87 & \textbf{61.29}\\
 \hline
\end{tabular}
\end{table*}

GP-TS can be trained in a few minutes on smaller datasets comprising only a few thousand images (12 Core-i7 CPU, 16 GB RAM), but slows significantly on extremely large datasets due to the GP's $\mathcal{O}(n^3)$ memory requirements. Ensembles and batched variational inference strategies remedy this to an extent, but deep learning approaches like FC-PWL, which can be trained more efficiently, are better suited to extremely large datasets, where sample efficiency is not required.

\subsection{Saliency}

\begin{figure}
\centering
\includegraphics[width=0.5\textwidth]{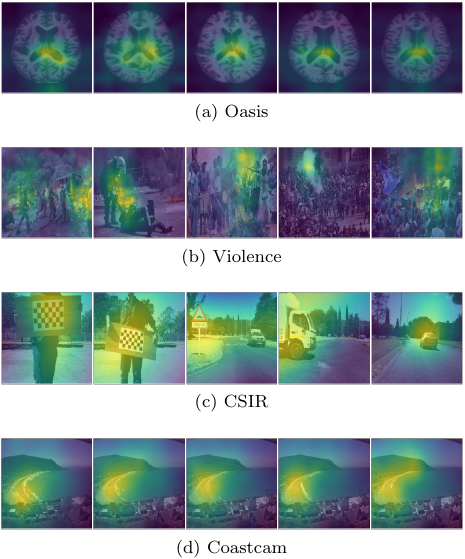}
\caption{Saliency maps show that the interest prediction model has identified content of interest to the end user. \label{saliency_overlay}}
\end{figure}

An occlusion-based sensitivity analysis technique \citep{zeiler2014visualizing} was applied to the trained models in order to investigate whether GP-TS is actually identifying image content of interest, or simply fitting to the data. Here, a blanking window is slid over the image, and the resultant change in predicted image interest measured at these blanked locations. Figure \ref{saliency_overlay} shows the 5 most interesting images in four test datasets, along with sensitivity maps.

It is clear that the model has learned to associate brain ventricles with interest in the Oasis dataset, while fire is highlighted in the violence dataset. In contrast, people and cars seem to be considered interesting in the CSIR set, while the coastline is associated with image interest for the Coastcam dataset.

\subsection{Storyboarding}
\label{story}

\begin{figure*}
\centering
\includegraphics[width=\textwidth]{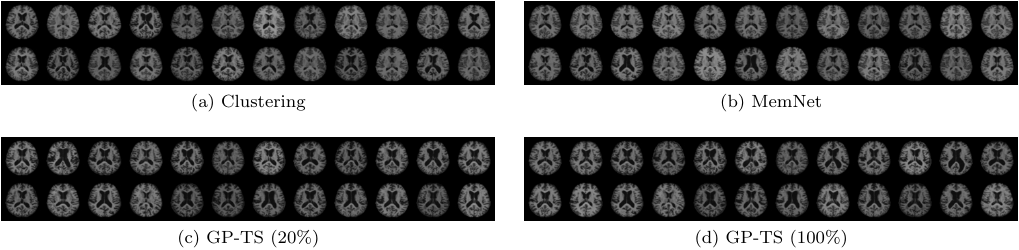}
\caption{OASIS dataset storyboards created using HAC, MemNet, and GP-TS show the range of brain scans in the dataset. Here, interesting images are those with reduced brain volume, typically indicated by enlarged central ventricles filled with fluid (coloured black).\label{OASIS_storyboarding}}
\end{figure*}

\begin{figure*}
\centering
\includegraphics[width=\textwidth]{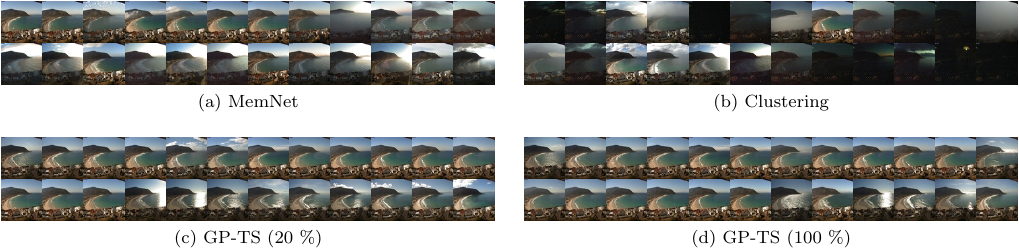}
\caption{The figure shows Coastcam dataset storyboards created using HAC, MemNet, and GP-TS. GP-TS storyboarding selects images with clearly differentiable wet and dry shoreline areas, and where the waves are in a backwash phase, in line with user preferences.\label{Coastcam_storyboarding}}
\end{figure*}

\begin{figure*}
\centering
\includegraphics[width=\textwidth]{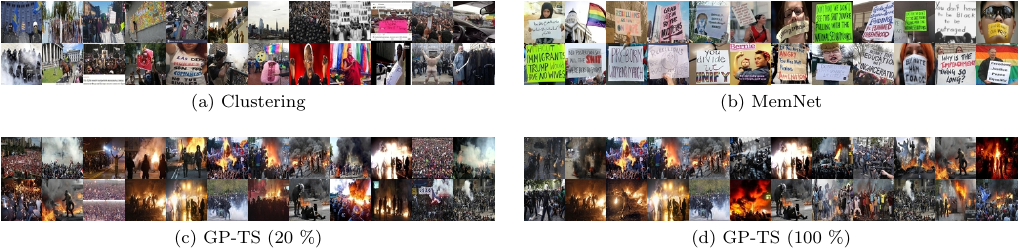}
\caption{Perceived violence dataset storyboards created using HAC, MemNet, and GP-TS highlight the differences between memorability and domain specific interest in violence.\label{Violence_storyboarding}}
\end{figure*}

\begin{figure*}
\centering
\includegraphics[width=\textwidth]{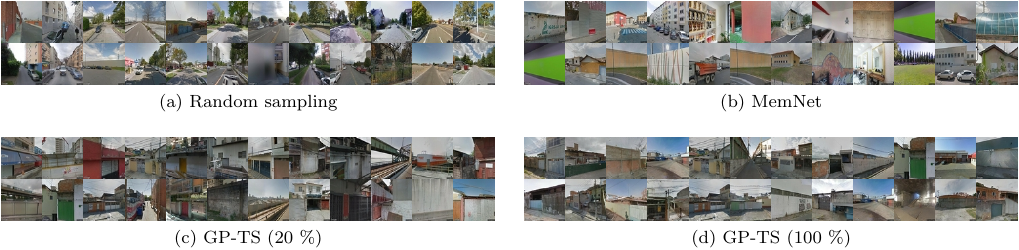}
\caption{Place Pulse 2.0 dataset storyboards created using random sampling, MemNet, and GP-TS highlight the differences between memorability and domain specific interest in street scene safety.\label{DLCity_storyboarding}}
\end{figure*}

Figure \ref{OASIS_storyboarding} shows 24-image storyboard summaries of the OASIS data set produced using GP-TS, MemNet and HAC. GP-TS storyboards were produced using both 100\% and 20\% of the available training data so as to highlight the rapid convergence to good interest estimates obtained using this approach. The GP-TS storyboard contains images likely to be of interest to an end user. In contrast, many commonly used storyboarding schemes lack the user-driven context of the proposed interest-based approach. Hierarchical agglomerative clustering produces a diverse set of images showing the range of healthy and unhealthy brains in the dataset, as the clustering rewards image dissimilarity, but many of the images produced are not of interest to an end-user. MemNet identifies a diverse range of images, but these fail to align with user preferences, while GP-TS has identified brains with enlarged ventricles as interesting. 

This is particularly noticeable if we consider the Coastcam storyboards shown in Figure \ref{Coastcam_storyboarding}. Here, HAC tends to show a diverse set of coastal conditions, which are certainly interesting to a general audience. MemNet restricts images in the storyboard to daylight images, but these storyboard images contrast significantly with the domain-specific interests of coastal scientists seeking to study soil erosion, as they fail to flag images with clearly distinguishable wet and dry sand regions.

The differences in storyboarding are even more stark when the Violence dataset is summarised using GP-TS, HAC and MemNet (Figure \ref{Violence_storyboarding}). HAC shows the broad range of images present in the dataset, MemNet seems to show a preference for signage, while GP-TS flags images with fire and fallen people as interesting. Similar results are visible when a storyboard of the Place Pulse 2.0 dataset is produced (Figure \ref{DLCity_storyboarding}). HAC is not used here, due to memory limitations.

\begin{table}
\centering
\caption{Number of interesting images per storyboard \label{tab_int}}
\begin{tabular}{|c|c|c|c|c|}
\hline
 & \bf{MemNet} & \bf{HAC} & \bf{GP-TS} & \bf{GP-TS}\\
 & & & \bf{20 \%} & \bf{100 \%}\\
\hline
\bf{OASIS} & 5 & 12 & 22 & 22\\
\hline
\bf{Violence} & 0 & 1 & 19 & 24\\
\hline
\bf{CSIR} & 17 & 15 & 23 & 24\\
\hline
\bf{Coastcam} & 9 & 1 & 24 & 24\\
\hline
\bf{Place Pulse} & 5 & - & 24 & 24\\
\hline
\end{tabular}
\end{table}

While it is clear that general purpose image summarising tools have their place, the storyboarding task above serves as an important reminder that in many instances, domain specific problems need to be solved. Here, image interest is often both task and problem dependent. This is highlighted by the simple count of interesting images present per storyboard provided in Table \ref{tab_int}.

\section{Memorability and Interest}
\label{memorability}

The relationship between image memorability and image interest warrants further investigation. Table \ref{mem_vs_interest} shows the Pearson correlation coefficients $\rho$ measured between memorability scores obtained using MemNet \citep{Khosla} and the domain-specific image interest predictions produced using GP-TS for each of the five test datasets, using all available pairwise comparisons for inference.  
\begin{table}
\centering
\small
\caption{Memorability vs Interest \label{mem_vs_interest}}
\begin{tabular}{|c|c|c|c|c|c|}
\hline
 & \bf{OASIS} & \bf{Violence} & \bf{CSIR}\\
\hline
 $\rho$ & -0.2 & -0.47 & 0.04\\
\hline
\hline
& \bf{Coastcam} & \bf{Place Pulse 2.0} &\\
\hline
$\rho$ & 0.7 & -0.32 &\\
\hline
\end{tabular}
\end{table}

Interestingly, memorability correlates the most with the interest scores obtained for the Coastcam database. This is potentially due to the fact that the coastal images of interest are typically captured in bright sunlight and are generally aesthetically pleasing, while there are a large number of dark images captured at night. There is a moderate negative correlation between image memorability and both the interests inferred from perceived violence measures and the street scene safety assessments in the Place Pulse 2.0 dataset. Similar results are obtained when measuring the correlation between memorability predictions and perceived violence scores directly ($\rho = -0.42$). This contrasts somewhat with the findings in \citep{Khosla}, which showed that there was little to no correlation between the aesthetic score of an image and its memorability, and that images that evoke anger and fear tend to be more memorable. 

It should be noted that the memorability predictions are made using a network that was trained using 60 000 images obtained from general image collections, and comprises both object-centric and scene-centric images, together with images of objects taken from unconventional angles, but was used in an entirely unsupervised manner here. As a result, it is possible that the memorability predictions are failing on the datasets investigated here.

\section{Conclusions}
\label{conclusions}

This paper has introduced a probabilistic pairwise ranking approach, GP-TS. Standard probabilistic ranking algorithms using pairwise comparisons like these typically require a large number of comparisons, but this work has shown that pairing these with a Gaussian process smoother dramatically reduces this number, by making use of similarities between image features extracted using a pre-trained convolutional neural network.

A primary benefit of GP-TS is that it produces a probability distribution over image interests. The uncertainty in these interest estimates can be used to select images to a present to a user for labelling, as part of an active learning process, but also to determine if sufficient data labelling has taken place. Existing optimisation-based ranking approaches do not allow for this, and tend to rely on large, labelled testing datasets to evaluate models. The probabilistic formulation allows for uncertainty resulting from unreliable comparisons that occurs when images appear visually similar to be captured. As a result, models trained using GP-TS are more suitable for rapid deployment, even if they do not necessarily perform well in all cases, because knowledge of when they fail to perform well is available. GP-TS significantly outperforms TS, a popular technique that is frequently used in pairwise image comparison studies because it provides reliable and stable results with confidence measures. The proposed approach is a drop-in replacement for TS that inherits its stable, probabilistic properties, while improving performance to the level of non-probabilistic state-of-the art approaches.

A number of inference strategies were considered for GP-TS, including variational inference under a fully Bayesian model, and decoupled inference using expectation propagation and a heteroscedastic Gaussian process. The latter proved most effective, with the decoupled inference strategy simplifying the inference process significantly, while improving prediction accuracy.

This work has also argued that image interest is often domain and task specific. A great deal of work has investigated general forms of image interest or memorability measures, but it is important to note that these measures are not always suitable for end-users. While there is indeed great value in collecting large scale datasets suitable for training general image interest and memorability scores, and this is extremely important for algorithm evaluation, practical deployments of efficient computer vision systems often require task specific algorithms that can be rapidly trained on small scale datasets.

\section*{Acknowledgements}
Thanks to Daniel Withey for valuable feedback, and Deon Sabatta and Christo Rautenbach for assistance with dataset collection.

\bibliographystyle{plainnat}      
\bibliography{references}   

\end{document}